# epiGPTope: A machine learning-based epitope generator and classifier

Natalia Flechas Manrique[§], Alberto Martínez[§], Elena López-Martínez, Luc Andrea, Román Orus, Aitor Manteca*, Aitziber L. Cortajarena*, Llorenç Espinosa-Portalés*

AUTHOR ADDRESS

Natalia Flechas Manrique – Multiverse Computing, Parque Científico y Tecnológico de Gipuzkoa, Paseo de Miramón 170, 20014 Donostia-San Sebastián, 20014 Spain

Alberto Martínez – Multiverse Computing, Parque Científico y Tecnológico de Gipuzkoa, Paseo de Miramón 170, 20014 Donostia-San Sebastián, 20014 Spain

Elena Lopez-Martinez - Centre for Cooperative Research in Biomaterials (CIC biomaGUNE), Basque Research and Technology Alliance (BRTA), Paseo de Miramón 194, Donostia-San Sebastián, 20014 Spain

Llorenç Espinosa-Portalés – Multiverse Computing, Parque Científico y Tecnológico de Gipuzkoa, Paseo de Miramón 170, 20014 Donostia-San Sebastián, 20014 Spain; https://orcid.org/0000-0002-7320-6769

Luc Andrea – Multiverse Computing, Multiverse Computing, WIPSE Paris-Saclay Enterprises 7, rue de la Croix Martre 91120 Palaiseau, France.

Román Orus – Multiverse Computing, Parque Científico y Tecnológico de Gipuzkoa, Paseo de Miramón 170, 20014 Donostia-San Sebastián, 20014 Spain; Donostia International Physics Center, Paseo Manuel de Lardizabal 4, E-20018 San Sebastián, Spain; Ikerbasque Foundation for Science, Maria Diaz de Haro 3, E-48013 Bilbao, Spain

Aitor Manteca - Centre for Cooperative Research in Biomaterialscurre (CIC biomaGUNE), Basque Research and Technology Alliance (BRTA), Paseo de Miramón 194, Donostia-San Sebastián, 20014 Spain; https://orcid.org/0000-0002-8650-0465

Aitziber L. Cortajarena - Centre for Cooperative Research in Biomaterials (CIC biomaGUNE), Basque Research and Technology Alliance (BRTA), Paseo de Miramón 194, Donostia-San Sebastián, 20014 Spain; IKERBASQUE, Basque Foundation for Science, Plaza Euskadi 5, 48009 Bilbao, Spain.

*Supporting Information Placeholder*

**ABSTRACT:** Epitopes are short antigenic peptide sequences which are recognized by antibodies or immune cell receptors. These are central to the development of immunotherapies, vaccines, and diagnostics. However, the rational design of synthetic epitope libraries is challenging due to the large combinatorial sequence space, $20^n$ combinations for linear epitopes of *n* amino acids, making screening and testing unfeasible, even with high throughput experimental techniques. In this study, we present a large language model, epiGPTope, pre-trained on protein data and specifically fine-tuned on linear epitopes, which for the first time can directly generate novel epitope-like sequences, which are found to possess statistical properties analogous to the ones of known epitopes. This generative approach can be used to prepare libraries of epitope candidate sequences. We further train statistical classifiers to predict whether an epitope sequence is of bacterial or viral origin, thus narrowing the candidate library and increasing the likelihood of identifying specific epitopes. We propose that such combination of generative and predictive models can be of assistance in epitope discovery. The approach uses only primary amino acid sequences of linear epitopes, bypassing the need for a geometric framework or hand-crafted features of the sequences. By developing a method to create biologically feasible sequences, we anticipate faster and more cost-effective generation and screening of synthetic epitopes, with relevant applications in the development of new biotechnologies.

## INTRODUCTION

The immune system protects the body from infections, but when dysregulated, it can lead to diseases such as autoimmunity[1], allergies[2], and immunodeficiencies[3]. Understanding how it works is key to improving vaccines, diagnostics, and immunotherapies[4,5]. A central part of the immune response involves the recognition of antigens to initiate a response. In this process short antigenic regions known as epitopes play a central role. Epitopes can be linear (**Figure 1**), defined only by the amino acid sequences, or conformational, depending on the 3D structure. This study focuses on linear epitopes derived from proteogenic antigens, where sequence analysis is essential. Identifying and modeling these sequences is a key challenge in immunology, particularly given the large size of the epitope sequence space and the limited availability of validated examples[6,7]. Even short peptides of *n* amino acids yield $20^n$ possible sequences, becoming unmanageable for experimental methods including high throughput screening technologies, such as droplet microfluidics[8,9] or phage display[10].

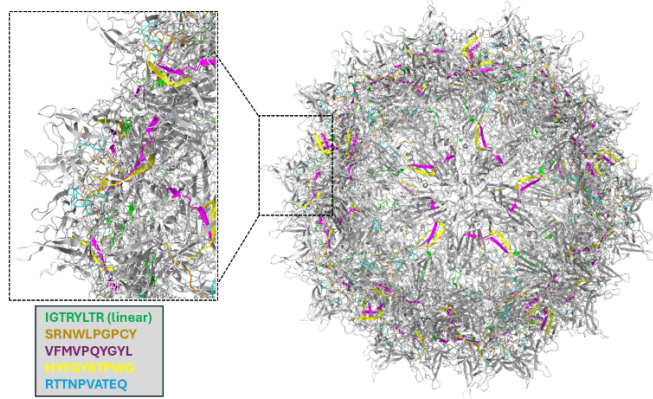

**Figure 1.** Structure of the adeno associated virus 2 (AAV2) (PDB ID: 1LP3). Colored regions show some of the epitopes already identified by various methodologies. The epitope highlighted in green is the only linear epitope, the rest are conformational.

.

While traditional work emphasized epitope mapping, recent advances have extended into modeling epitope–antibody recognition using deep learning[11] and structural bioinformatics[12], offering powerful tools for predictive profiling. To reduce the number of candidate sequences, several approaches aim to find patterns or statistical properties that make certain epitopes more plausible candidates. In this sense, natural Language Processing (NLP) have shown promising results in identifying underlying patterns in epitope sequences[13]. NLP is a rich and rapidly progressing field that provides tools able to find and operate with the logical structures that emerge from the concatenation of letters and words. Recent progress has been particularly notorious thanks to the introduction of the transformer architecture and the self-attention mechanisms[14], which are, nowadays, at the heart of most large language models (LLMs) implementations, such as BERT[15], T5[16], GPT[17–19] or LLaMa[20]. These LLMs can perform linguistic tasks beyond the reach of previous computational tools and previous deep learning / machine learning architectures, such as support vector machines, RNNs or LSTMs[21]. Like natural language, amino acid sequences are information-complete, they encode the necessary information for biological function. This is especially evident in short, linear sequences like epitopes, but also applies to longer sequences that adopt specific 3D structures in their native, low-energy states. Inspired by this analogy, NLP tools have been applied to proteins, including generative models like ProtGPT2[22], and predictive models such as ProtBERT[23] and epiBERTope[24]. The latter is trained on epitope data, and is able to classify sequences as potential linear or conformational epitopes. Motivated by these advances, the approach presented here introduces a fully generative component in a computational epitope discovery pipeline. This novel generative model, epiGPTope, is able to generate de novo epitope-like amino acid sequences that can be further filtered by classification models.

In this work, we apply and extend NLP methods for epitope design and engineering, with the goal of generating synthetic epitope libraries. By training machine learning (ML) models on natural epitopes, we aim to propose candidates with higher binding potential in laboratory screening, reducing experimental costs and accelerating discovery. Our pipeline consists of two components: a generative module and a classifier. The generative model, epiGPTope, was developed by fine-tuning ProtGPT2 on curated natural epitope sequences from the Immune Epitope Database (IEDB) [25,26] using unsupervised learning. Thus, epiGPTope learns the statistical properties of linear epitopes and encodes them on a probability distribution. This allows to generate new epitope-like amino acid sequences by sampling from the distribution in a sequential manner. To evaluate and filter the generated sequences, we trained a family of classifiers using positive and negative examples from IEDB [25,26], whose effectivity when filtering a library was evaluated by its Positive Likelihood Ratio (LR+). We focused specifically on viral and bacterial epitopes, as these represent the largest and most consistently annotated subsets in the IEDB, providing a rich and diverse training source with reliable labels. The complete workflow, including data preparation, model training, hyperparameter tuning, epitope generation, and library analysis, is presented here. All models, including epiGPTope and the classifiers, are accessible via Singularity Computational Biology by Multiverse Computing at: https://epitope-library-generation.singularity-quantum.com. Access information and trial can be provided upon request.

**METHODS**

**Data preparation**. Amino acid sequences were retrieved from the IEDB[26]. An initial filtering process was performed based on the host (*homo sapiens*), the type of assay (T cell, B cell or MHC) and the structure of the epitope (linear). Thereafter, the dataset was further filtered such that it excluded sequences with more than 11 residues. Duplicated sequences were removed. In addition, the format of the pre-training dataset was emulated. The total number of sequences used for fine-tuning was 504611. Our previous work shows the statistical characteristics of these epitopes[13]. While the whole dataset is used to train epiGPTope, the accompanying classifiers are trained on eight subsets restricted filtered according to the organism (bacterial or viral) and the assay (Tcell, Bcell, MHC or all combined), as well as on negative data.

**Model fine-tuning and sequence generation.** To generate new sequences, we have followed a fine-tuning approach, which has been successful in other artificial intelligence fields such as computer vision and natural language processing[27]. Fine-tuning is a form of transfer learning in which a language model is pre-trained on a self-supervised task, such as language modeling. Afterwards, the pre-trained model is fine-tuned by adjusting its weights on the target task's data[28]. This approach is based on pre-training tasks such as causal language modeling (predict a word given its left-side context) and seems deceptively simple, although their correct resolution hinges on the general understanding of the structure that underpins the sequences[15,29]. Such understanding is then used as a starting point for specialized tasks during fine-tuning.

In our case, we chose protGPT2[22] as the model to be fine-tuned. ProtGPT2 shares the architecture of the original GPT2 model created for natural language [30], as shown in **Figure 2**. It has a total of 738 million parameters, arranged in 6 layers with a model dimensionality of 1280. In this case, the pre-training was done on a causal modeling objective over UniRef50 (version 2021 04), a clustering of UniProt at 50% identity[31]. A language model assigns a probability to a sentence or sequence of tokens $W$ of length $n$. In the case of an autoregressive model such as the one we used, this probability is further defined as the product of the individual probabilities of each word $w_i$ given its left-side context, as shown in equation 1.

$$p(W) = \prod_{i=1}^{n} p(w_i \| w_{<i})$$ (1)

The model seeks to minimize the negative log-likelihood (equation 2) over the entire dataset.

$$l_{CLM} = -\sum_{k=1}^{D} \log p_\theta \left(w_i^k \| w_{<i}^k\right)$$ (2)

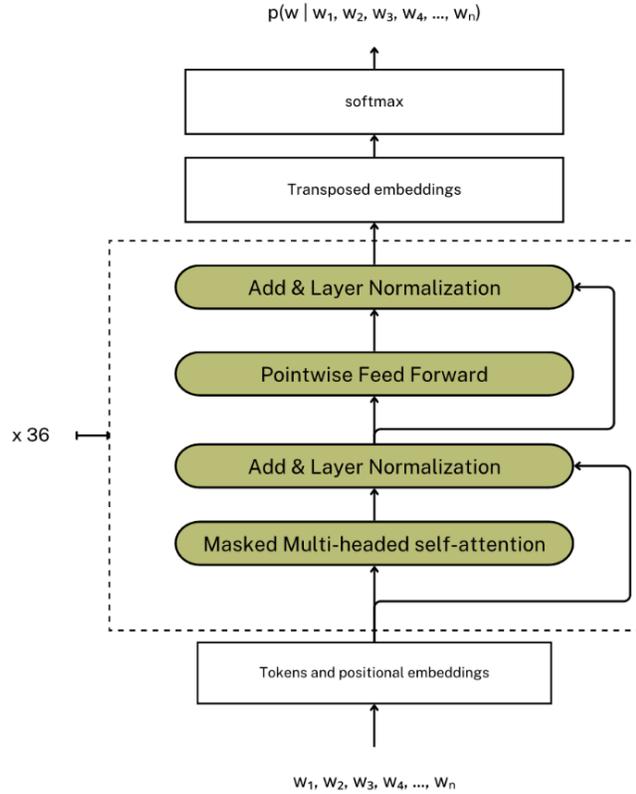

**Figure 2.** Architecture of the GPT2-like large model, which was designed for natural language processing, and is also used by ProtGPT2 and epiGPTope.

We performed the fine-tuning by minimizing the same objective over the filtered IEDB sequences. An AWS EC2 G5.2xlarge instance was used, which is a high-performance computer instance equipped with an NVIDIA A10G Tensor Core GPU, offering 24 GB of video memory, 80 ray tracing cores, 320 tensor cores, and a total throughput of 250 TOPS. After stress testing, it was determined that optimal batch size should be 48. We experimented with different set-ups of hyperparameters, which are depicted in **Table 1**.

**Table 1. Fine-tuning hyperparameters.** We have tested 5 different models to fine-tune our system with different combinations of learning rates, epochs, decays, and GPU types.

| Model | Learning Rate | Epochs | Weight Decay | GPU type |
|---|---|---|---|---|
| 1 | 0.001 | 15 | 0.001 | NVIDIA A10G |
| 2 | 0.01 | 15 | 0.01 | NVIDIA A10G |
| 3 | 0.001 | 30 | 0.01 | NVIDIA A10G |
| 4 | 0.001 | 30 | 0.01 | Tesla T4 |
| 5 | 0.01 | 30 | 0.01 | Tesla T4 |

*Relative entropy.* We calculated the relative entropy between the observed probability distribution of amino acids at each residue position and that of the human proteome (equation 3). For two discrete probability distributions $P$ and $Q$ defined on sample space X, the relative entropy is defined as:

$$S_{P|Q} = \sum_{x \in X} P(x) \log \left(\frac{P(x)}{Q(x)}\right)$$ (3)

which can be understood as a measure of how different both distributions are.

*Global propensity of amino acids* at each residue position (equation 4). The global propensity of amino acid $X$ at position $y$ in a sample with estimated probability $p_{X_y}$ and a reference probability in the human proteome $q_{X_y}$ is defined as the ratio

$$propensity_{X,y} = \frac{p_{X,y}}{q_{X,y}}$$ (4)

*Shannon entropy of the amino acid* distribution at each residue position (equation 5). For a discrete probability distribution $P$ on a sample space $X$, it is defined as

$$S_P = -\sum_{x \in X}^{P}(x) \log P(x)$$ (5)

which can be understood as an inverse measure of information or difference with an information-less random uniform distribution (**Figure 4d** and **Supp. Fig. 3.**).

*Perplexity* of a sequence is a measure of its compatibility with a discrete probability distribution. The lower its value, the better. For a tokenized sequence $X = (x_0, x_1, ..., x_t)$, it is defined as follows:

$$PPL(x) = \exp\left\{-\frac{1}{t}\sum_{i=0}^{t} \log p_\theta(x_i|x_{<i})\right\}$$ (6)

**Classification.** After the generation of epitopes, we sought a method to quantify the efficacy of the generated sequences. Beyond comparing the statistical analyses between epitopes from established datasets and those generated in our study, we coded a binary classifier specifically designed to differentiate between epitopes and non-epitope sequences, for both bacterial and viral epitopes.

Both the generative and classification models are part of a ML pipeline designed to obtain a library of synthetic epitopes, which can then be tailored to a particular experimental application, as shown in **Figure 3**. In a nutshell, the generative step obtains a general-purpose library by taking samples from the complex probability distribution of the epitope data, learnt by epiGPTope from as many data as possible, while the classification step aims to filter the library to increase the likelihood of success for a particular application. Here, the classifiers filter candidates by epitope organism of origin, viral vs. bacterial, but the same strategy could be extended to other properties to train other application-specific classifiers.

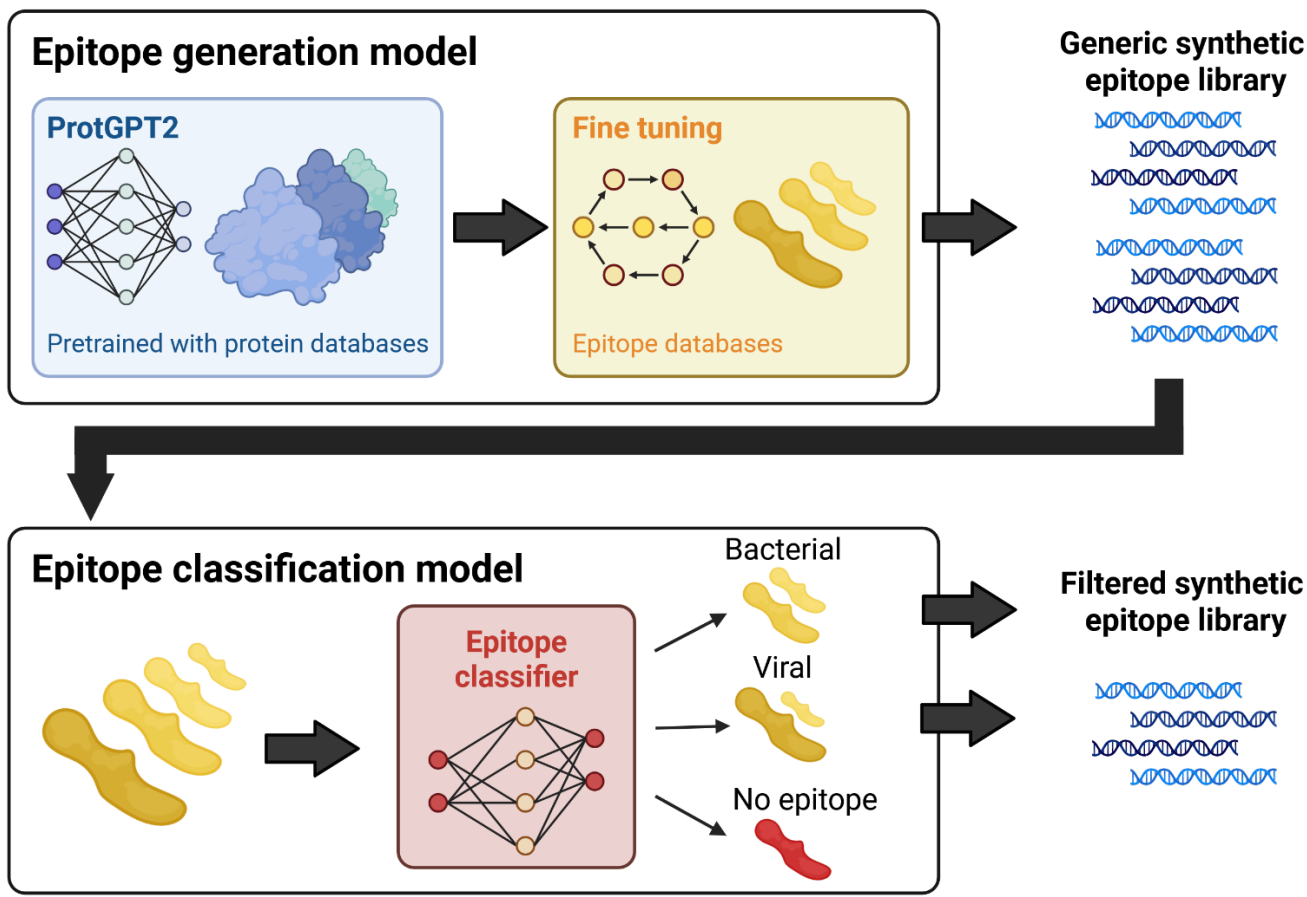

**Figure 3.** Flowchart depicting the design of a system that generates a library of synthetic epitopes and tailors it to a particular experimental application. The generation step is leverages the general-purpose epiGPTope model. The classification step can apply any desired filtered, powered by an ML model. Here, as an example, we trained a family of classifiers that recognize linear epitopes of various organisms and obtained in different types of assays. This figure was created using www.biorender.com.

**RESULTS**

**Generation of sequences.** The fine-tuned language model has been used to generate 192,222 distinct synthetic epitope sequences. The statistical properties of the generation data, the experimentation and selection of the hyperparameters used in generation and the details on the classification task are discussed in this section. We adjusted 2 hyperparameters for sequence generation: repetition penalty [1.2, 2, 3] and temperature [2, 0.5]. In machine learning, particularly in training neural networks and generating probabilistic outputs[32], temperature is a parameter that modulates the level of uncertainty or randomness in a model's predictions. Adjusting the temperature value affects the diversity of generated samples, with higher values promoting a more explorative and varied output, while lower values result in more focused and deterministic predictions. This parameter is often applied in tasks like text generation to control the trade-off between exploration and exploitation during the sampling process. As for repetition penalty, it can be defined as a method used to address the challenge of generating repetitive sequences in natural language generation models, such as those based on recurrent neural networks[33] (RNNs) or transformers. It involves penalizing the model for producing repeated tokens or phrases in its generated output. By incorporating repetition penalty during training or decoding, the model is incentivized to generate more diverse and coherent outputs, reducing the likelihood of redundant or monotonous patterns.

Building upon prior research focused on the statistical analysis of natural epitopes[13], we conducted a comprehensive statistical examination of synthetic epitopes. This analysis included an evaluation of sequence length distribution (**Figure 4a**) alongside a series of additional statistical computations. The results of the distribution analysis indicate that the most prevalent epitope size falls within the range of 7 to 9 amino acids. This finding is consistent with previously reported analyses of the IEDB, further supporting the notion that synthetic epitopes exhibit size characteristics similar to those observed in naturally occurring epitopes. These results provide valuable insights into epitope length preferences. Global propensity analyses (**Figure 4b, Figure 4c, Supp. Fig. 1. and Supp. Fig. 2**) reveal that the prevalence of aromatic residues in epitopes is particularly notable in the final position of these short sequences, contrasting with previously reported aromatic residue propensities in conformational epitopes, but in accordance with our previous results[7]. This discrepancy may be attributed to potential pi-stacking interactions between the aromatic residues within the epitopes and the antibody, which could play a role in molecular recognition[34]. Another notable observation is the low frequency of cysteines across all epitope positions. Cysteines can form covalent disulfide bonds,[35] which are stable and irreversible under physiological conditions. Since antibody−antigen recognition is a high-affinity yet reversible interaction, the presence of cysteines may be disfavor, and their nature may interfere with the transient nature of the binding.

Estimating probability distributions and calculating Shannon entropy are generally complex tasks due to the inherent variability and potential biases in finite datasets. However, in this study, we assume that the sample size is sufficiently large to ensure that the observed frequencies serve as reliable approximations of the true underlying probabilities. This assumption allows for a more robust statistical analysis while minimizing the impact of sampling errors. Previous studies have employed similar approaches[36] to derive meaningful insights from empirical frequency distributions. Our results (**Figure 4d and Spp. Fig. 3**) reveal that the information shared between the different residues that contain the algorithm is neglectable. No correlations are observed between pairs of residue positions for a given sequence length, provided that the number of generated sequences for that length is sufficiently large. This is a positive indication that the model does not introduce spurious correlations between amino acids. However, for sequence lengths with fewer generated sequences, sizable mutual information is detected, likely due to the high uncertainty in estimating this quantity from a small sample size. Moreover, in **Figure 5**, the training losses of different experiments are shown. In machine learning, an epoch refers to one complete pass through the entire training dataset by the model. Since training data is often divided into smaller batches, an epoch consists of multiple iterations where the model updates its weights after each batch. Training typically involves multiple epochs to allow the model to learn meaningful patterns and minimize error over time. It is important to note that we chose the model of experiment 1 to generate the final sequences. This is because we do not rely only on the losses, but also on the statistical characteristics depicted in **Figure 4**.

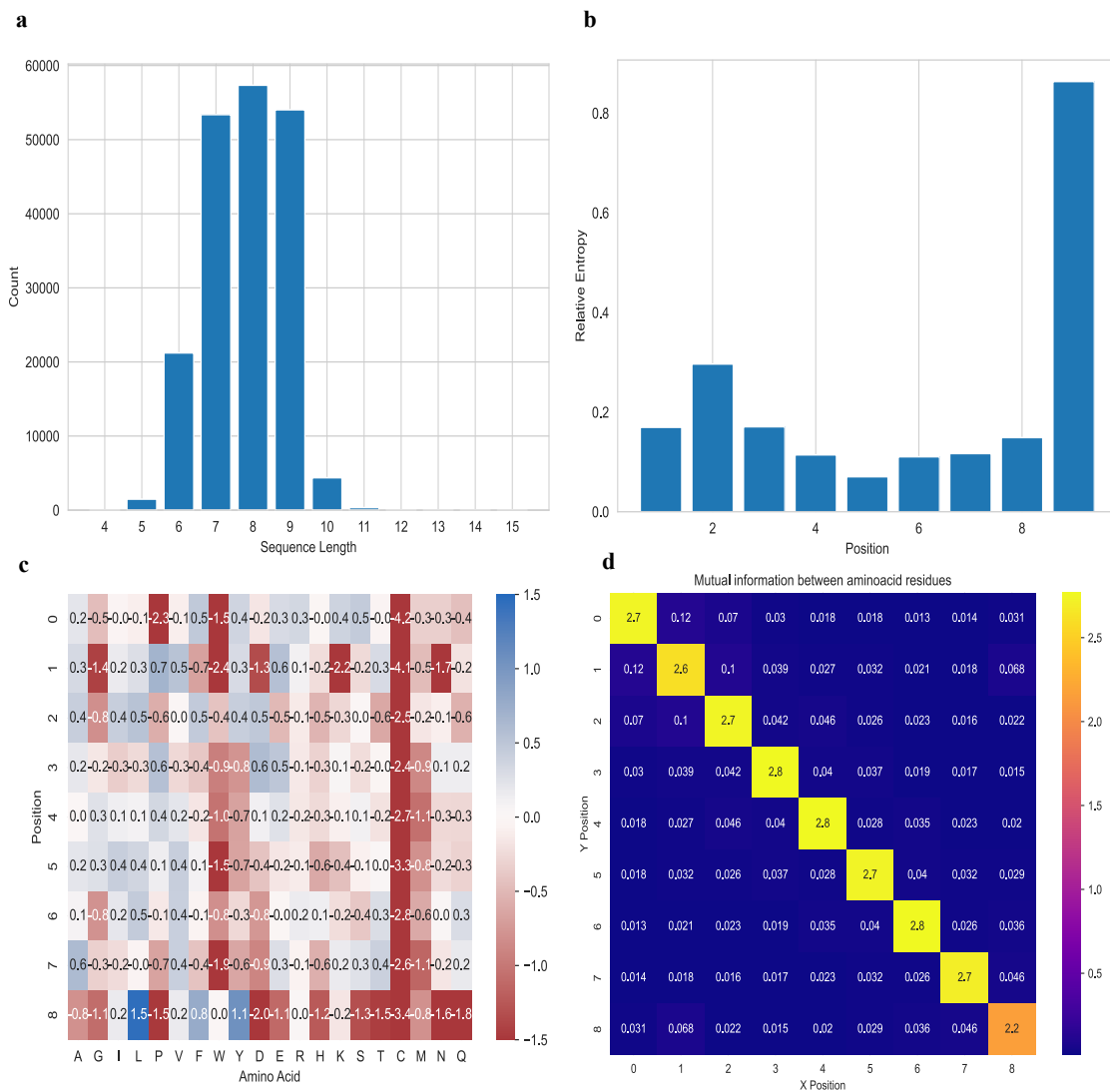

**Figure 4.** a) Distribution of sequence lengths in the dataset, showing a sharp peak at 8–9 residues, which reflects the dominant length of linear epitopes. b) Relative entropy (Kullback–Leibler divergence) for each position in 9-residue sequences, indicating positional conservation. Higher values suggest lower variability (greater conservation), with position 9 showing the strongest bias. c) Heatmap of relative amino acid propensities by position for sequences with 9 residues. Values indicate log-odds scores (red: overrepresented; blue: underrepresented). The color intensity reflects deviation from background frequency (standard amino acid usage), helping identify dominant or excluded residues at each position. d) Pairwise mutual information (MI) between positions in 9-residue sequences, quantifying positional dependence. The matrix shows MI values between each position pair (X vs. Y). Higher values (yellow) suggest co-dependence, potentially reflecting structural or binding constraints, while lower values (dark blue) indicate independent variation.

We decided to further experiment with generation hyperparameters. For this purpose, we generated around 100k sequences with different repetition penalties: the default value was 1.2, and we additionally experimented with 2 and 3. We obtained the perplexities as described in the Hugging face documentation[37] and performed statistical testing. The mean score for repetition penalty 1.2 was 5414.087, while the mean for repetition penalty 2 was 5207.757. We compared both generations' mean perplexities with the Mann-Whitney test. The Mann-Whitney U statistic was 48867288108.0, and the associated p-value was $8.7e^{-52}$. The null hypothesis was rejected, indicating a significant difference between the two conditions. For repetition penalty 3, the mean score was 5270.926. Similar to the previous case, we compared this condition with the default value (repetition penalty = 1.2). The Mann-Whitney U statistic was 48867288108.0, and the p-value was 8.7e-52. Consequently, the null hypothesis was rejected, underscoring a significant difference associated with Repetition Penalty 3. Finally, a comparison test between the generation with repetition penalty 2 and 3 yielded: a Mann-Whitney U Statistic of $4.685*10^9$ and a P-value of 0.757. The null hypothesis is not rejected, indicating no significant difference between the two conditions. All things considered; we chose a repetition penalty of 2 for the final generation of sequences as it is the one that presents the lowest mean perplexity value. Small temperature experiments (10k sequences) were produced for different temperature values [2, 1, 0.5], with repetition penalty = 2. Statistical tests were not performed in this case. Statistical properties were analyzed, and they appeared similar. For this reason, we chose a temperature value of 1.

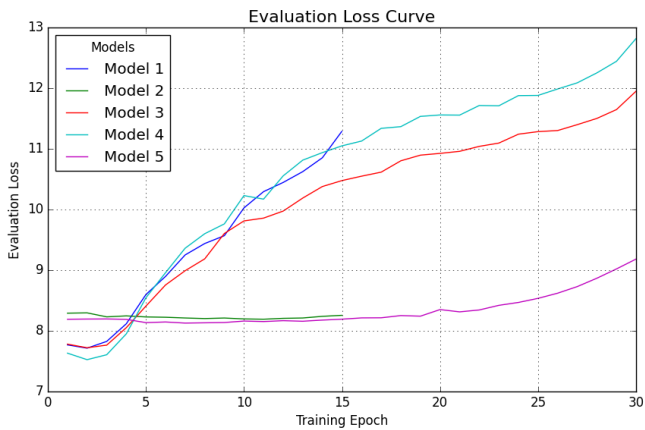

**Figure 5.** Evaluation loss curves across five model training experiments. This plot shows the evaluation loss over 30 training epochs for five different model configurations (Models 1–5). Models 1, 3, and 4 display a clear minimum loss early in training, typically around epoch 2–5, after which loss increases steadily—a sign of overfitting. In contrast, Models 2 and 5 exhibit flat evaluation loss curves, with little to no improvement over the training period, suggesting poor learning dynamics or failure to converge. These results help identify configurations that are prone to overfitting versus those that may require adjustment in architecture or hyperparameters to enable learning.

**Classification.** We present a family of classifiers trained on bacterial and viral data, which classify amino acid sequences as epitopes or non-epitopes. They are trained on subdivisions of the IEDB database as described in the data preparation subsection. We gather the performance metrics of the ensemble-based classifiers in **Table 2**, and those of the LLM-based classifiers in **Table 3**. Our first approach was to use an ensemble classifier based on XG-Boost library[38] build on top of vector embeddings. To that end, we compared ProtGPT2, ProtBERT, and a bare BERT model pre-trained on natural language for baseline. We extended our research to include embeddings derived from the fine-tuned version of ProtGPT2 that we had generated for the generation task. There was the expectation that these fine-tuned embeddings would offer higher accuracy compared to the embeddings generated by the baseline ProtGPT2 model. However, our experimental findings revealed that we obtained comparable results, indicating that the fine-tuning did not yield the expected improvement in the predictive performance of embeddings, which is not entirely surprising given that the fine-tuning was not tailored to the classification task, but rather to the generative one. As a matter of fact, while ProtGPT2 provided the best embedding for bacterial epitopes, the bare BERT baseline turned to outperform other embeddings for viral ones.

**Table 2.** Metrics of the ensemble-based family of classifiers: $F_1$ score, accuracy, recall, precision, and LR+.

| Model | Organism | F1 | Accuracy | Recall | Precision | LR+ |
|---|---|---|---|---|---|---|
| Ensemble | Bacterial | 0.331 | 0.493 | 0.795 | 0.209 | 1.41 |
| Ensemble | Viral | 0.559 | 0.485 | 0.818 | 0.425 | 1.13 |

**Table 3.** Metrics of the LLM-based family of classifiers: $F_1$ score, area under the ROC curve, area under the PR curve, and LR+, which gauges the efficiency of the classifier as a filter. Among all of them, those trained with data from MHC binding assays show the best performance.

| Model | Organism | Assay | F1 | ROC AUC | PR AUC | LR+ |
|---|---|---|---|---|---|---|
| ProtBERT | Bacterial | TCell | 0.49 | 0.686 | 0.417 | 16.247 |
| ProtBERT | Bacterial | BCell | 0.519 | 0.762 | 0.47 | 3.658 |
| ProtBERT | Bacterial | MHC | 0.869 | 0.642 | 0.818 | 1.187 |
| ProtBERT | Viral | T-cell | 0.565 | 0.676 | 0.537 | 1.841 |
| ProtBERT | Viral | B-cell | 0.601 | 0.918 | 0.524 | 6.697 |
| ProtBERT | Viral | MHC | 0.852 | 0.719 | 0.812 | 1.465 |
| ProtGPT2 | Bacterial | T-cell | 0.481 | 0.734 | 0.453 | 15.716 |
| ProtGPT2 | Bacterial | B-cell | 0.453 | 0.71 | 0.465 | 2.930 |
| ProtGPT2 | Bacterial | MHC | 0.865 | 0.667 | 0.83 | 1.173 |
| ProtGPT2 | Viral | T-cell | 0.583 | 0.711 | 0.595 | 1.667 |
| ProtGPT2 | Viral | B-cell | 0.587 | 0.916 | 0.546 | 6.949 |
| ProtGPT2 | Viral | MHC | 0.846 | 0.767 | 0.872 | 1.391 |
| ProtGPT2 | Bacterial | T-cell | 0.481 | 0.734 | 0.453 | 10.054 |
| ProtBERT | Bacterial | All | 0.5 | 0.716 | 0.432 | 11.779 |
| ProtGPT2 | Bacterial | All | 0.502 | 0.739 | 0.45 | 3.439 |
| ProtBERT | Viral | All | 0.607 | 0.912 | 0.61 | 3.253 |
| ProtGPT2 | Viral | All | 0.58 | 0.899 | 0.561 | 16.247 |

Additionally, we also tried different methods to get an embedding of a sequence. LLMs give us the embedding of each token in their context, but a sequence is usually composed of several tokens. This computing was carried out by three different ways: (i) keeping the rightmost last hidden state, (ii) adding all last hidden states, and (iii) including all last hidden states with the rightmost being assigned more weight. Option (ii) delivered the best performance. During the experiments, we realized that if we take 100 vector elements, we get almost the same results as the whole embedding vector. So, we created an ensemble of classifiers, an *ensemble of ensembles*, whose inputs are 100 vector elements, and make decisions by biased majority voting. This ensemble allows us to bias the classifier and achieve a high recall, which is our target, as we aim to remove as few true epitopes as possible. In essence, the bias is a parameter that determines the weight assigned to each model's output. If the output of the individual model is positive, the weight equals the value of the bias, while if the output of the individual model is negative, the weight is 1.

To understand how these classifiers help filtering and improving an epitope candidate dataset, we assumed that the proportion of positive and negative examples is given by the quantities $p_+$ and $p_-$. After applying the classifier and retaining only the resulting positive sequences, we used conditional probabilities to evaluate the transition from the ratio in the dataset to the ratio after classification (equation 7).

$$\frac{p_+}{p_-} \to \frac{TPR}{FPR} \cdot \frac{p_+}{p_-} = (LR+) \cdot \frac{p_+}{p_-} \quad (7)$$

The improvement on the library brought by the filter is quantified by Positive Likelihood Ratio $LR+$, a common metric to quantify the success of a diagnostic test in medicine, defined as the quotient of the True Positive Ratio (TPR) and the False Positive Ratio (FPR). Such quantity has the advantage of being independent of the class imbalance of the library prior to its filtering by the classifier.

**Dimensional reduction of epitope data.** Our initial phase involved a comprehensive study of the data and their inherent properties. Each epitope in our dataset is represented by a substantial 1280-dimensional vector, coming from the vector embeddings of ProtGPT2. This led us to the question of whether we could effectively reduce this dimensionality without losing crucial information. Given the intricate nature of the ProtGPT2 embeddings and their lack of interpretable features, we excluded the possibility of manual dimensional reduction. Therefore, we explored the feasibility of employing automatic dimension reduction techniques such as Principal Component Analysis (PCA) and Uniform Manifold Approximation and Projection (UMAP). We show the results of performing PCA up to the 2 or 3 most significant features in **Supp. Fig. 4.** PCA is a statistical method used for reducing the dimensionality of a dataset while preserving as much variance as possible. It transforms the original features into new uncorrelated variables called principal components, which are ranked based on the amount of variance they capture. PCA is commonly used for data compression, noise reduction, and as a preprocessing step in machine learning to improve efficiency and visualization. UMAP is a dimensionality reduction technique designed for visualizing and analyzing high-dimensional data in a lower-dimensional space (commonly 2D or 3D). It is faster than t-SNE, preserves more of the global and local structure, and is based on mathematical principles from graph theory and topology. UMAP is widely used in clustering, gene expression analysis, and neural network feature visualization. No patterns or potential clustering are identified, a signal of the high complexity of the data despite consisting of linear sequences of ~ 10 amino acids.

## DISCUSSION

Overall, these results are a successful application of NLP techniques to synthetic epitope design. In particular, they show how to obtain new sequences with a generative language model, as shown by the statistical measures and their striking similarity to those in the training set and previously reported in the literature[13]. In this sense, epiGPTope can correctly capture features present in the training set, that are qualitatively understood to characterize epitopes, (e.g. such as a large relative entropy and small Shannon entropy in the last residue position or a lower propensity of cysteines). Furthermore, no correlations are found between pairs of residue positions for a given sequence length, provided the number of generated sequences for that length is sufficiently large. This is a positive sign that the model is not adding spurious correlations between amino acids. Sizable mutual information appears for those sequence lengths with few generated sequences due to a high uncertainty in the estimate of this quantity for a small sample size. Little is known about the nature of these complex patterns, as it is not analyzed in previous work on natural epitopes and is hidden within the non-interpretable parameters of epiGPTope in the case of synthetic ones.

A more careful inspection of the positive and negative epitope data in the IEDB reveals that many statistical sequence features commonly associated with epitopes—such as low cysteine frequency or elevated relative entropy at certain positions—are also found in negative examples. This observation suggests that these features may not, in themselves, define epitopes, but may instead reflect selection biases in experimental design. For instance, candidate sequences may be chosen to exclude cysteines (to avoid potential complications associated to disulfide bonds) or to enrich for sequence variability at terminal positions. To disentangle these effects, we introduce the correlation between relative entropy and positional conservation. Relative entropy measures deviation from background amino acid frequencies and can indicate evolutionary constraint, but it may also reflect artificial filtering criteria used during data curation. Positions with high relative entropy often correspond to more conserved regions, but not always—especially when certain residues are intentionally under- or over-sampled in experimental design. Therefore, it is essential to distinguish between features that reflect biological relevance (e.g., conservation across pathogens) and those that arise from systematic biases in the way epitope candidates are selected or reported. This distinction is subtle but critical: if such features influence the training data without truly reflecting the nature of the sequence, then models may learn to replicate selection biases rather than genuine immunological principles. Incorporating this understanding is essential for building models that generalize beyond the quirks of existing datasets and better reflect the biology of antibody-epitope interaction.

Since IEDB data is compiled from numerous independent experiments, we could not determine if there exist common procedures to select candidate sequences for experimental testing. Second, it is important to recognize that many of the so-called negative examples in epitope prediction datasets may not be true negatives, but rather epitopes for antibodies not included in the experimental assay. This label noise introduces ambiguity and may partly explain why certain models still achieve good performance on global metrics: they tend to capture general patterns of immunogenicity, rather than true antibody-specific binding. For example, widely used tools such as BepiPred and epiBERTope rely on amino acid properties or pre-trained language models to identify residues likely to be part of epitopes. While effective in broad classification tasks, these models are typically not designed to distinguish whether a given sequence will bind a specific antibody. As a result, they can identify sequences that look "epitope-like"—according to shared physicochemical or structural features—but may not actually interact with the antibody of interest. This highlights a core limitation in current approaches: success in epitope identification does not necessarily translate into precision in epitope detection. Bridging this gap will require the development of models that integrate epitope data, binding assay results, or structural constraints, allowing for the prediction of epitope–antibody compatibility rather than epitope potential alone. This is particularly crucial for therapeutic antibody development, vaccine design, and diagnostics, where target specificity is essential.

Moreover, structure-based methods have emerged to address the limitations of sequence-only epitope predictors by incorporating spatial context and even antibody-specific information. Tools like PEASE[39], which integrates antibody sequence with antigen structure to predict antibody-specific epitopes, represent a key advance Similarly, ElliPro[40] and PEPITO[41] utilize antigen 3D structures to identify conformational epitopes based on solvent accessibility and residue exposure. More recent approaches, including DeepInterAware[42] and AbAgIntPre[43], use deep learning with sequence or structure representations of both antibody and antigen to predict binding interfaces, reporting ROC-AUC values around 0.8  Tools like PEASE bridge this gap to some extent by being antibody-aware, but their performance still depends heavily on structural data availability. In contrast, our LLM-based approach is designed to directly capture antibody-specific binding signatures from sequence data, enabling precise epitope predictions without requiring expensive structural information. Integrating such specificity-aware sequence models with structurally informed methods represents a promising direction for developing next-generation tools with both accessibility and precision.

The results on the classifiers presented in Section 2 hint at the statistical difference between positive and negative examples of IEDB. These are a complement to the generator model and are meant to filter a generic epitope library to increase the likelihood of interaction with a paratope. In the experimental framework, the selection of a dataset becomes pivotal, considering its impact on the performance and capabilities of the binary classifier. Thus, it is necessary to work with an adjustable classifier, so that the bias can be adjusted to the dataset or the experimental requirements. Ensemble classifiers with various language embeddings were designed to that end. Given the better performance of the bare BERT embeddings for viral epitopes, which was unexpected, we wondered whether the embeddings were really capturing information of biological relevance. This motivated us to dive deeper in the classifiers and build a family of fine-tuned LLMs based on ProtGPT2 and ProtBERT. To achieve it, we selected several data subsets from IEDB with viral and bacterial epitopes, and further subdivisions according to whether the epitopes were identified in T-Cell, B-Cell or Major Histocompatibility Complex (MHC) binding assays for a deeper investigation. These were used to train specific classifiers, which, unlike the generator, are aware of both positive and negative epitope examples for each category. Fine-tuned versions of ProtGPT2 and ProtBERT were obtained for these tasks. We see the addition of ProtBERT for this task due to its architecture being more amenable for classification tasks.

The performance of the LLM-based classifiers varied notably depending on the source of the training data, as summarized in Table 3. Models trained on MHC binding assay data consistently showed superior performance across all evaluation metrics compared to those trained on broader or less experimentally grounded datasets. This underscores a critical insight: the biological specificity and experimental reliability of the training data are just as important as model complexity in achieving accurate epitope prediction. These findings have several implications. First, they reinforce the idea that model performance in immunological tasks is tightly coupled to data provenance and curation. High-throughput datasets with well-characterized biological endpoints, such as MHC binding assays, provide a stronger foundation for machine learning than datasets derived from indirect or purely computational annotation. Second, the improved performance of models trained on these data types suggests that future efforts should prioritize integrating experimentally validated immune data—even if smaller in volume—over larger but noisier datasets. Finally, this result opens the door for transferring learning approaches, where models are pre-trained on general epitope data but fine-tuned on MHC-specific or high-quality subsets to improve task-specific accuracy. In the broader context of synthetic epitope design and immunodiagnostics, these results point to a promising strategy: building predictive tools that explicitly leverage biologically rich, assay-derived data can lead to more reliable predictions of antibody-epitope interactions, increasing the chances of successful experimental validation downstream.

We find that both the bacterial and viral models can tell apart epitopes from non-epitopes, which is consistent with previous work on the matter[44]. Even though this does not exclude the possibility that existing classifiers learn non-generalizable features of the dataset, i.e., that they fine-tune to the training plus test sets, current dataset sizes make us confident of the existence of distinct epitope characteristics that are captured by the classifiers. We find the metrics obtained by our classifiers to be highly satisfactory, especially considering we have trained on the largest dataset size to our knowledge. Furthermore, their high LR+ values prove their usefulness when leveraged as filters of a general-purpose library that can tailor it to wet for lab experiments. The difficulty of improving classifier performance for epitope data can be seen from the inability of dimensional reduction techniques such as PCA or UMAP to provide a minimal clustering of the data (See **Supp. Fig 4**). Both calculations fail to recognize any clustering event.

There are a few improvements we identified that might be implemented in future work. The epiGPTope model is quite large (1.5 billion parameters and about 9GB in memory). Given that linear epitope sequences consist of tens of amino acids, while full-length proteins span hundreds of residues, we expect the model to be potentially compressed to remove unnecessary information on long-range correlations between residues. Quantum-inspired Tensor Networks (TN) are particularly suitable for this task[45]. This could help to further test more hyperparameter combinations, which could lead to performance improvements. Currently, the computational expense of this task makes this exploration difficult. Furthermore, we wonder about the possibility of further fine-tuning ProtGPT2 to be able to perform additional tasks, such as tailoring the generated sequences to specific antibody targets. This is an exciting possibility, albeit challenging due to the epitope-paratope relationship being not 1-1 and the lack of enough data on its multivalued nature, i.e., data rich enough to match each of many epitopes to many paratopes in turn. Finally, it would be interesting to mutate given sequences in a way that introduces variability but keeps their nature as epitope candidates.

## SUMMARY

The use of ML algorithms and AI is revolutionizing every field it meets. Biotechnology is not an exception, and, in fact, one may argue that bioinformatics is an early adopter of many of the new technologies being developed. In this work, we explored the application of LLMs to epitope design. Specifically, we trained a generative model and two classifiers, which leverage transfer learning from ProtGPT2, a LLM trained on abundant protein sequence data. Our results show that it is possible to generate and filter synthetic amino acid sequences as viable epitope candidates, and classify their origin (viral or bacterial). This classification step increases the likelihood of generating application-relevant epitopes. We anticipate that these computationally derived candidates will support experimental efforts to discover new epitope sequences that target specific antibodies in immunological research.

## ASSOCIATED CONTENT

### Supporting Information

The Supporting Information is available free of charge on the ACS Publications website.

Supp. Fig. 1. Entropies of the generated epitopes per residue position from $n = 6$ to $n = 13$.
Supp. Fig. 2. Global propensities of the generated epitopes per residue position from $n = 6$ to $n = 13$.
Supp. Fig. 3. Shannon entropies of the generated epitopes per residue position from $n = 6$ to $n = 13$.
Supp. Fig. 4. Global Propensities for the training data.
Supp. Fig. 5. Dimensional reduction of the vector embeddings of bacterial and viral datasets.

## AUTHOR INFORMATION

### Corresponding Author

Aitor Manteca: amanteca@cicbiomagune.es; Aitziber L. Cortajarena: alcortajarena@cicbiomagune.es; Llorenç Espinosa-Portalés: llorenc.espinosa@multiversecomputing.com

### Author Contributions
§These authors contributed equally.

### Notes
All the authors declare competing financial interest regarding the patent called Neoepitope- A machine learning-based epitope generator and classifier with European Application No. 24383487.6.
.

## ACKNOWLEDGMENT

A.M. is financed by grant 2022-FELL-000011-01 funded by Gipuzkoa Fellows Program (Diputación Foral de Gipuzkoa), AECC Grant INVES258597MANT-ImmunoDrop and grant "EPINPOC" cofunded by AECT Euroregion New Aquitaine-Navarra-Basque Country. A.L.C. acknowledges financial support from Grants, PID-2022-137977OB-I00, and PDC2021-120957-I00 funded by MCIN/AEI/10.13039/501100011033 and by the "European Union NextGenerationEU/PRTR". A.M. and A.L.C. also acknowledges financial support from Diputación Foral de Gipuzkoa grant 2023-QUAN-000023-0, 2023-KA44-000007-01 and 2024-QUAN-000017-01.

## REFERENCES

1. Wang L, Wang F, Gershwin ME. Human autoimmune diseases: a comprehensive update. *Journal of internal medicine*. 2015;278(4):369-395.
2. Crameri R. Allergy diagnosis, allergen repertoires, and their implications for allergen-specific immunotherapy. *Immunology and Allergy Clinics*. 2006;26(2):179-189.
3. Ballow M. Primary immunodeficiency disorders: antibody deficiency. *Journal of Allergy and Clinical Immunology*. 2002;109(4):581-591.
4. Palatnik-de-Sousa CB, Soares I da S, Rosa DS. Epitope discovery and synthetic vaccine design. *Frontiers in immunology*. 2018;9:347379.
5. Lin MJ, Svensson-Arvelund J, Lubitz GS, et al. Cancer vaccines: the next immunotherapy frontier. *Nature cancer*. 2022;3(8):911-926.
6. Sela-Culang I, Ofran Y, Peters B. Antibody specific epitope prediction—emergence of a new paradigm. *Current opinion in virology*. 2015;11:98-102.
7. Akbar R, Robert PA, Pavlović M, et al. A compact vocabulary of paratope-epitope interactions enables predictability of antibody-antigen binding. *Cell Reports*. 2021;34(11). doi:10.1016/j.celrep.2021.108856
8. Manteca A, Gadea A, Van Assche D, et al. Directed Evolution in Drops: Molecular Aspects and Applications. *ACS Synthetic Biology*. 10(11):2772-2783. doi:10.1021/acssynbio.1c00313
9. Sieskind R, Cortajarena AL, Manteca A. Cell-Free Production Systems in Droplet Microfluidics. *Cell-free Macromolecular Synthesis*. Published online 2023:91-127.
10. Jaroszewicz W, Morcinek-Orłowska J, Pierzynowska K, Gaffke L, Węgrzyn G. Phage display and other peptide display technologies. *FEMS Microbiology Reviews*. 2022;46(2):fuab052.
11. Abanades B, Wong WK, Boyles F, Georges G, Bujotzek A, Deane CM. ImmuneBuilder: Deep-Learning models for predicting the structures of immune proteins. *Communications Biology*. 2023;6(1):575.
12. Dunbar J, Krawczyk K, Leem J, et al. SAbDab: the structural antibody database. *Nucleic acids research*. 2014;42(D1):D1140-D1146.
13. Lopez-Martinez E, Manteca A, Ferruz N, Cortajarena AL. Statistical Analysis and Tokenization of Epitopes to Construct Artificial Neoepitope Libraries. *ACS Synthetic Biology*. Published online 2023.
14. Vaswani A, Shazeer N, Parmar N, et al. Attention is all you need. *Advances in neural information processing systems*. 2017;30.
15. Devlin J, Chang MW, Lee K, Toutanova K. BERT: Pre-training of Deep Bidirectional Transformers for Language Understanding. Published online May 24, 2019. doi:10.48550/arXiv.1810.04805
16. Raffel C, Shazeer N, Roberts A, et al. Exploring the Limits of Transfer Learning with a Unified Text-to-Text Transformer. *Journal of Machine Learning Research*. 2020;21(140):1-67.
17. Radford A, Narasimhan K, Salimans T, Sutskever I. Improving Language Understanding by Generative Pre-Training.
18. Brown TB, Mann B, Ryder N, et al. Language Models are Few-Shot Learners. Published online July 22, 2020. doi:10.48550/arXiv.2005.14165
19. OpenAI, Achiam J, Adler S, et al. GPT-4 Technical Report. Published online March 4, 2024. doi:10.48550/arXiv.2303.08774
20. Touvron H, Lavril T, Izacard G, et al. LLaMA: Open and Efficient Foundation Language Models. Published online February 27, 2023. doi:10.48550/arXiv.2302.13971
21. Pertseva M, Gao B, Neumeier D, Yermanos A, Reddy ST. Applications of machine and deep learning in adaptive immunity. *Annual Review of Chemical and Biomolecular Engineering*. 2021;12:39-62.
22. Ferruz N, Schmidt S, Höcker B. ProtGPT2 is a deep unsupervised language model for protein design. *Nature communications*. 2022;13(1):4348.
23. Elnaggar A, Heinzinger M, Dallago C, et al. ProtTrans: Towards Cracking the Language of Life's Code Through Self-Supervised Learning. Published online May 4, 2021:2020.07.12.199554. doi:10.1101/2020.07.12.199554
24. Park M, Seo S woo, Park E, Kim J. EpiBERTope: a sequence-based pre-trained BERT model improves linear and structural epitope prediction by learning long-distance protein interactions effectively. Published online March 2, 2022:2022.02.27.481241. doi:10.1101/2022.02.27.481241
25. Vita R, Overton JA, Greenbaum JA, et al. The immune epitope database (IEDB) 3.0. *Nucleic acids research*. 2015;43(D1):D405-D412.
26. Vita R, Mahajan S, Overton JA, et al. The immune epitope database (IEDB): 2018 update. *Nucleic acids research*. 2019;47(D1):D339-D343.
27. Zhuang F, Qi Z, Duan K, et al. A comprehensive survey on transfer learning. *Proceedings of the IEEE*. 2020;109(1):43-76.
28. Basile P, Musacchio E, Polignano M, Siciliani L, Fiameni G, Semeraro G. LLaMAntino: LLaMA 2 Models for Effective Text Generation in Italian Language. Published online December 15, 2023. doi:10.48550/arXiv.2312.09993
29. Sutskever I, Vinyals O, Le QV. Sequence to Sequence Learning with Neural Networks. Published online December 14, 2014. doi:10.48550/arXiv.1409.3215

30. Radford A, Wu J, Child R, Luan D, Amodei D, Sutskever I. Language models are unsupervised multitask learners. *OpenAI blog*. 2019;1(8):9.
31. UniProt: the universal protein knowledgebase in 2023. *Nucleic acids research*. 2023;51(D1):D523-D531.
32. Specht DF. Probabilistic neural networks. *Neural networks*. 1990;3(1):109-118.
33. Caterini AL, Chang DE, Caterini AL, Chang DE. Recurrent neural networks. *Deep neural networks in a mathematical framework*. Published online 2018:59-79.
34. Arzhanik V, Svistunova D, Koliasnikov O, Egorov AM. Interaction of antibodies with aromatic ligands: The role of π-stacking. In: *Journal of Bioinformatics and Computational Biology*. Vol 8. ; 2010:471-483. doi:10.1142/S0219720010004835
35. Cremlyn RJ, Cremlyn RJW. *An Introduction to Organosulfur Chemistry*. John Wiley & Sons; 1996.
36. Paninski L. Estimation of Entropy and Mutual Information. *Neural Computation*. 2003;15(6):1191-1253. doi:10.1162/089976603321780272
37. Face H. Transformers documentation. *URL: https://huggingface co/docs/transformers/index*. Published online 2024.
38. Chen T, Guestrin C. XGBoost: A Scalable Tree Boosting System. In: *Proceedings of the 22nd ACM SIGKDD International Conference on Knowledge Discovery and Data Mining*. ; 2016:785-794. doi:10.1145/2939672.2939785
39. Sela-Culang I, Ashkenazi S, Peters B, Ofran Y. PEASE: predicting B-cell epitopes utilizing antibody sequence. *Bioinformatics*. 2015;31(8):1313-1315.
40. Ponomarenko J, Bui HH, Li W, et al. ElliPro: a new structure-based tool for the prediction of antibody epitopes. *BMC bioinformatics*. 2008;9:1-8.
41. Sweredoski MJ, Baldi P. PEPITO: improved discontinuous B-cell epitope prediction using multiple distance thresholds and half sphere exposure. *Bioinformatics*. 2008;24(12):1459-1460.
42. Xia Y, Wang Z, Huang F, et al. DeepInterAware: Deep Interaction Interface-Aware Network for Improving Antigen-Antibody Interaction Prediction from Sequence Data. *Advanced Science*. Published online 2025:2412533.
43. Huang Y, Zhang Z, Zhou Y. AbAgIntPre: A deep learning method for predicting antibody-antigen interactions based on sequence information. *Frontiers in Immunology*. 2022;13:1053617.
44. Bahai A, Asgari E, Mofrad MR, Kloetgen A, McHardy AC. EpitopeVec: linear epitope prediction using deep protein sequence embeddings. *Bioinformatics*. 2021;37(23):4517-4525.
45. Tomut A, Jahromi SS, Sarkar A, et al. CompactifAI: Extreme Compression of Large Language Models using Quantum-Inspired Tensor Networks. Published online May 13, 2024. doi:10.48550/arXiv.2401.14109

Insert Table of Contents artwork here

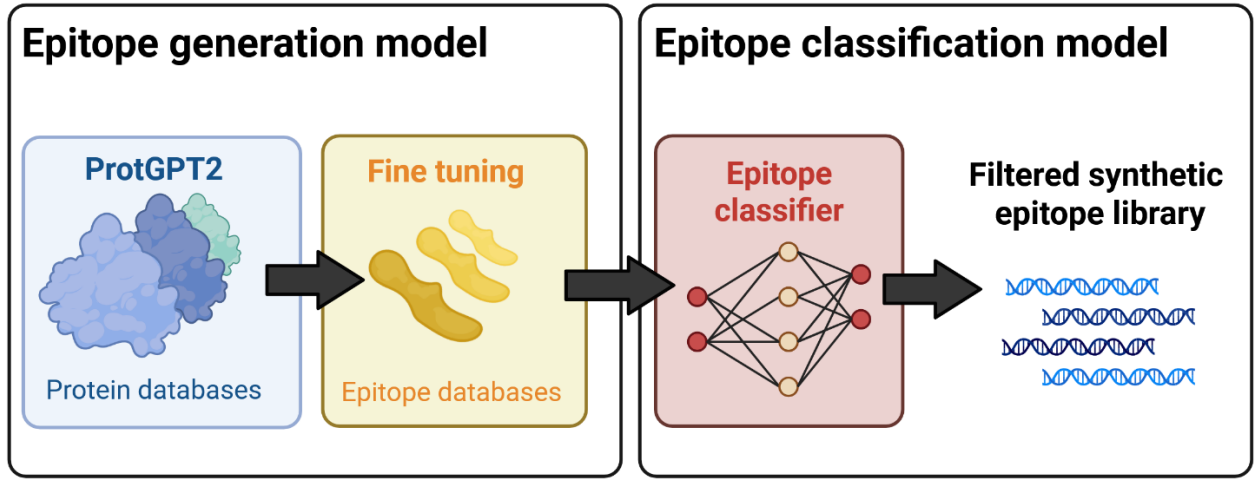

Supplementary information for:

# epiGPTope: A machine learning-based epitope generator and classifier

Natalia Flechas Manrique[§], Alberto Martínez[§], Elena López-Martínez, Luc Andrea, Román Orus, Aitor Manteca*, Aitziber L. Cortajarena*, Llorenç Espinosa-Portalés*

AUTHOR ADDRESS

Natalia Flechas Manrique – Multiverse Computing, Parque Científico y Tecnológico de Gipuzkoa, Paseo de Miramón 170, 20014 Donostia-San Sebastián, 20014 Spain

Alberto Martínez – Multiverse Computing, Parque Científico y Tecnológico de Gipuzkoa, Paseo de Miramón 170, 20014 Donostia-San Sebastián, 20014 Spain

Elena Lopez-Martinez - Centre for Cooperative Research in Biomaterials (CIC biomaGUNE), Basque Research and Technology Alliance (BRTA), Paseo de Miramón 194, Donostia-San Sebastián, 20014 Spain

Llorenç Espinosa-Portalés – Multiverse Computing, Parque Científico y Tecnológico de Gipuzkoa, Paseo de Miramón 170, 20014 Donostia-San Sebastián, 20014 Spain; https://orcid.org/0000-0002-7320-6769

Luc Andrea – Multiverse Computing, Multiverse Computing, WIPSE Paris-Saclay Enterprises 7, rue de la Croix Martre 91120 Palaiseau, France.

Román Orus – Multiverse Computing, Parque Científico y Tecnológico de Gipuzkoa, Paseo de Miramón 170, 20014 Donostia-San Sebastián, 20014 Spain; Donostia International Physics Center, Paseo Manuel de Lardizabal 4, E-20018 San Sebastián, Spain; Ikerbasque Foundation for Science, Maria Diaz de Haro 3, E-48013 Bilbao, Spain

Aitor Manteca - Centre for Cooperative Research in Biomaterialscurre (CIC biomaGUNE), Basque Research and Technology Alliance (BRTA), Paseo de Miramón 194, Donostia-San Sebastián, 20014 Spain; https://orcid.org/0000-0002-8650-0465

Aitziber L. Cortajarena - Centre for Cooperative Research in Biomaterials (CIC biomaGUNE), Basque Research and Technology Alliance (BRTA), Paseo de Miramón 194, Donostia-San Sebastián, 20014 Spain; IKERBASQUE, Basque Foundation for Science, Plaza Euskadi 5, 48009 Bilbao, Spain.

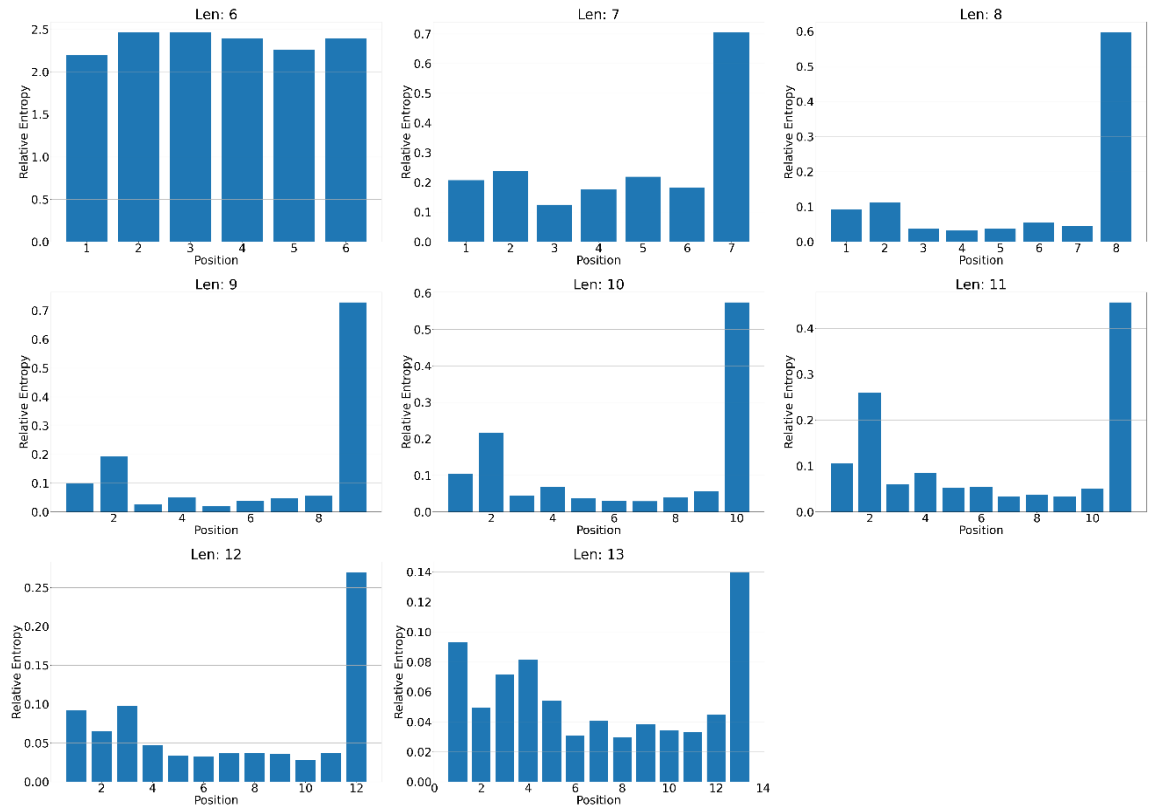

**Supplementary Figure 1. Positional distribution of relative entropy across epitope sequences of varying lengths (n = 6 to n = 13).** Bar plots show the relative entropy at each position within epitopes of a given length, indicating positional conservation. For shorter epitopes (e.g., n = 6), entropy is uniformly high across positions, suggesting consistent conservation. In contrast, longer epitopes (n ≥ 7) exhibit elevated entropy primarily at the terminal positions, especially at the final residue, highlighting position-specific conservation patterns.

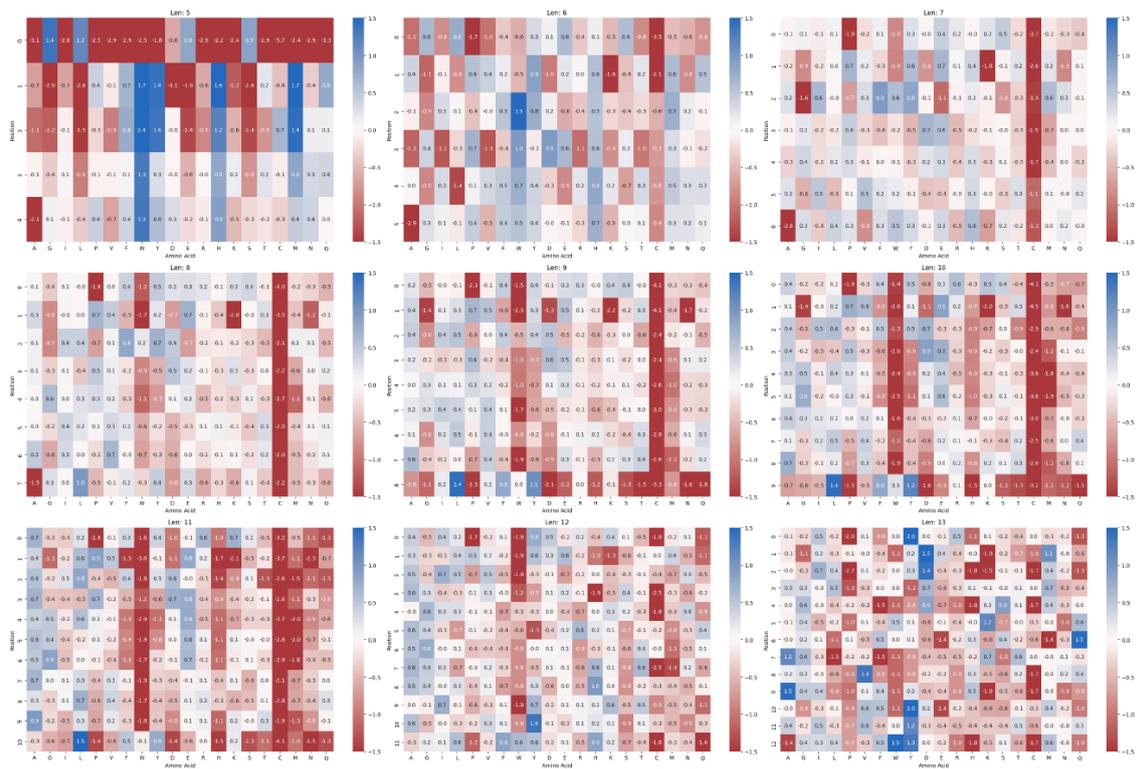

**Supplementary Figure 2. Positional amino acid enrichment across epitope sequences of varying lengths (n = 5 to n = 14).** Heatmaps display normalized log-odds scores for amino acid frequencies at each position within epitopes of fixed lengths. Rows represent amino acids, and columns indicate positions within the epitope. Red and blue shading indicate enrichment and depletion, respectively, relative to background frequencies. Distinct positional preferences are observed, with certain residues consistently enriched or depleted at specific locations, suggesting conserved sequence features and functional constraints across epitope lengths.

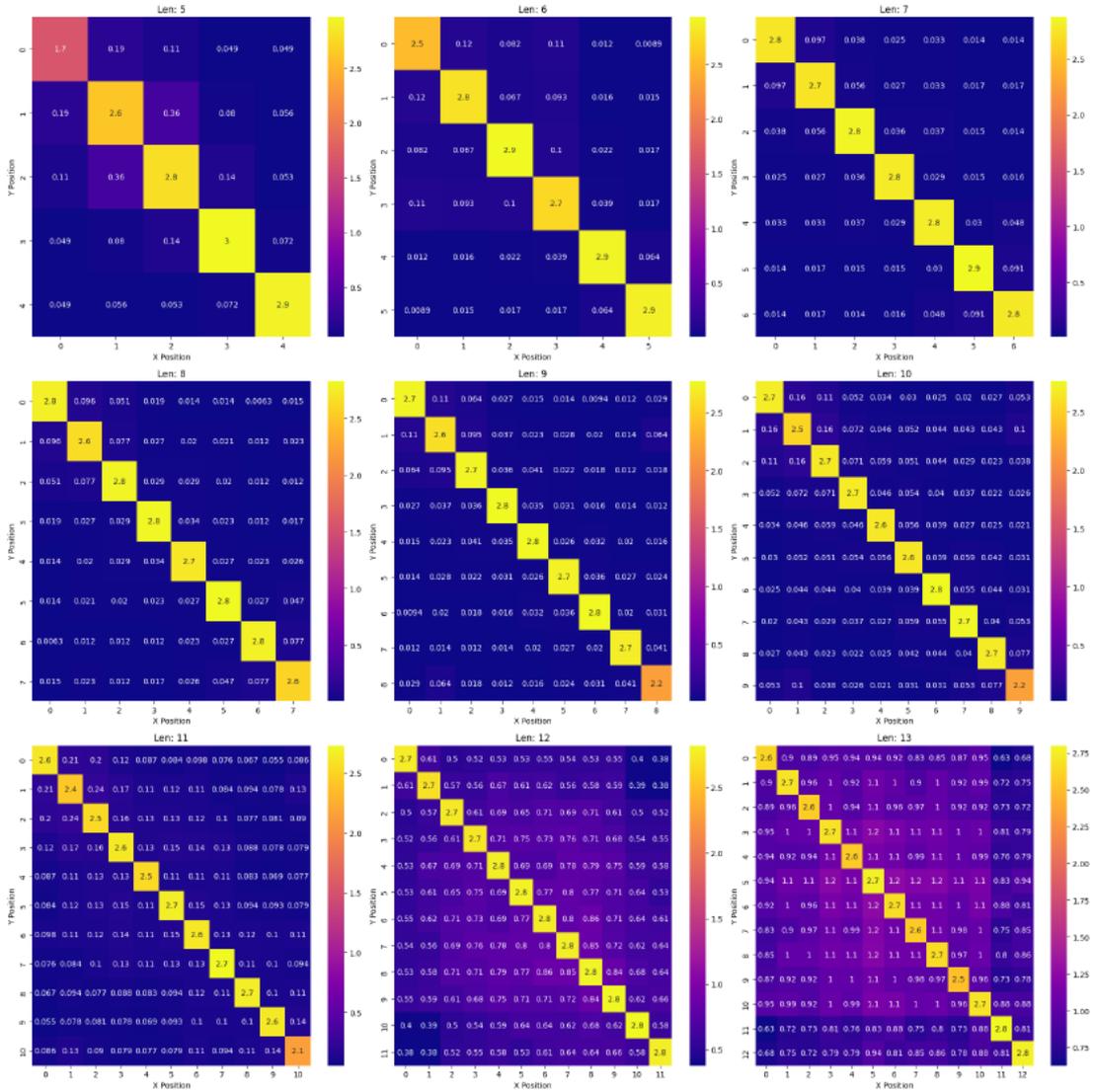

**Supplementary Figure 3. Shannon entropy matrices for epitope sequences of lengths n = 5 to n = 14.** Heatmaps represent the Shannon entropy between positional pairs within epitope sequences of specified lengths. Each cell indicates the entropy value between position i (Y-axis) and position j (X-axis), quantifying the uncertainty or variability in amino acid combinations at those positions. Diagonal elements reflect single-position entropy, while off-diagonal values capture joint variability between positions. Higher entropy values (yellow) suggest greater sequence diversity, whereas lower values (blue/purple) indicate conservation or positional dependency.

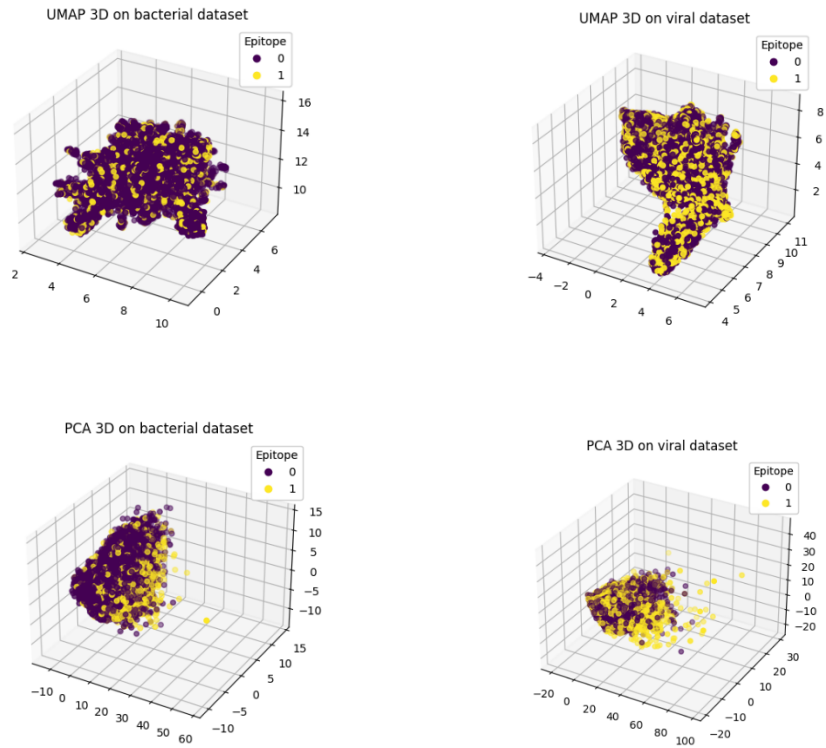

**Supplementary Figure 4.** UMAP and PCA 3D representations of bacterial and viral datasets, illustrating the distribution and clustering patterns of sequences within these datasets. These visualizations provide insights into the structural and compositional differences between bacterial and viral epitopes, highlighting potential distinctions in their sequence properties.